# DETECTION AND CLASSIFICATION OF VIEWER AGE RANGE SMART SIGNS AT TV BROADCAST[*]

Baran Tander[1] Atilla Özmen[2] and Murat Başkan[3]

[1]Vocational School of Technical Sciences, Communication Electronics Program,
Kadir Has University, 34590, Selimpaşa, Silivri, İstanbul, Turkey
`tander@khas.edu.tr`
[2]Faculty of Engineering and Natural Sciences, Electrical – Electronics Engineering
Department, Kadir Has University, 34083, Cibali, Fatih, İstanbul, Turkey
`aozmen@khas.edu.tr`
[3]Vocational School of Technical Sciences, Mechatronics Program,
Kadir Has University, 34590, Selimpaşa, Silivri, İstanbul, Turkey
*baskan@khas.edu.tr*

## ABSTRACT

*In this paper, the identification and classification of "Viewer Age Range Smart Signs", designed by the Radio and Television Supreme Council of Turkey, to give age range information for the TV viewers, are realized. Therefore, the automatic detection at the broadcast will be possible, enabling the manufacturing of TV receivers which are sensible to these signs. The most important step at this process is the pattern recognition. Since the symbols that must be identified are circular, various circle detection techniques can be employed. In our study, first, two different circle segmentation methods for still images are analyzed, their advantages and drawbacks are discussed. A popular neural network structure called Multilayer Perceptron is employed for the classification. Afterwards, the same procedures are carried out for streaming video. All of the steps depicted above are realized on a standard PC.*

## 1. INTRODUCTION

"Smart Signs" shown in Figure 1 are designed by the Radio and Television Supreme Council of Turkey (RTUK), in order to provide the following information of a particular program at Turkish TV channels [1]:

  1-) The content of the program,
  2-) The appropriate viewer age range

Just like the Motion Picture Association of America (MPAA) movie rating system [2].

Project No. BAP0907001

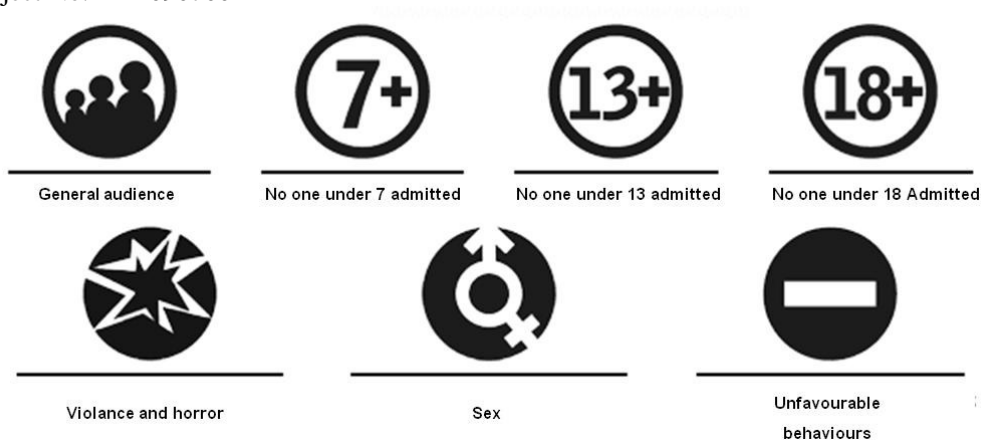

Figure 1.Smart signs.

The viewer age range smart signs (7+, 13+, 18+) among them are symbols which are automatically classified at our work, appear approximately for 10 seconds at the upper-left or upper-right corners of the broadcast at the beginning of the programs and after the commercial breaks.As seen from the figure, since all of the symbols are circular, utilizing a circle detection method will be suitable for the segmentation.

The organization of the paper is as follows:Firstly, a pre-processing scheme on the frames taken from the broadcast, in order to decrease the segmentation time is presented. Secondly, two circle detection algorithms, namely Circle Hough Transform (CHT) and Circle Estimation (CE) are discussed. The feature extraction process of the segmented signs to introduce them to a Multilayer Perceptron (MLP) Neural Network for the classification of the viewer age range is given afterwards. All of theseprocedures are shown in Figure 2.

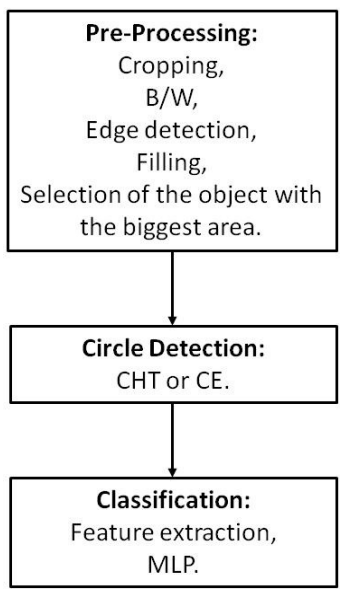

Figure 2.The whole detection and classification process.

Finaly, simulations are carried out to observe the performance and speed of the whole process both for still images and streaming video.

## 2. PRE – PROCESSING

A pre-processing including the following steps, has to be carried out before the circle detection in order to increase the segmentation speed of the viewer age range smart signs, to make the real-time operation of the system possible:

1-) <u>Cropping:</u> Since the viewer age range smart signs are located at the upper-left or upper-right corners of the frames, the cropping of those two will enable one to operate in relatively small areas which will ease the circle detection. The cropped area which doesn't include the smart sign will be eliminated at the classification,

2-) <u>Conversion to black and white:</u> Converting the cropped areas to black and white will be quite useful before the edge detection step, since the processor will have to deal with only grayscaled images which will reduce the computational burden,

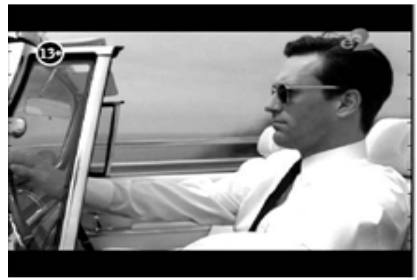

(a)

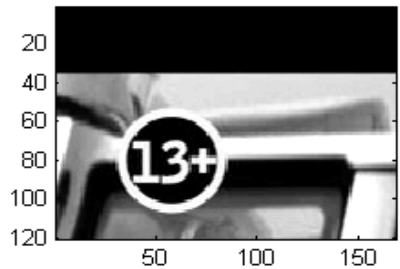
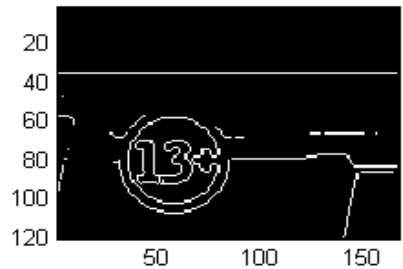

(b)        (c)

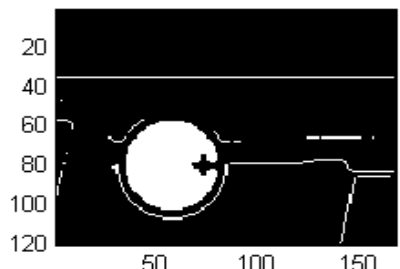
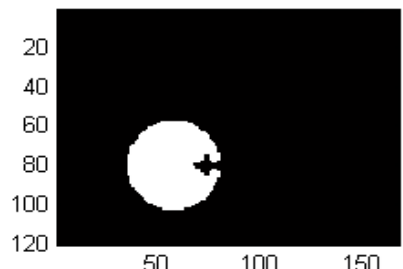

(d)        (e)

Figure 3. Pre-processing of the frame: (a) Original image captured from the TV broadcast, (b) Cropped B/W image, (c) Edge detection, (d) Closed object filling, (e) Selection of the filled object with the biggest area.

3-)Edge detection: A Sobel filter can be employed for the edge detection process [3] to convert the cropped areas to a binary image,

4-)Filling: Filling the closed objects will enable us to select the potential viewer age range smart signs at the cropped areas,

5-)Selection of the object having the biggest area: It is obvious that, generally the viewer age range smart sign whose center will be determined is the filled object with the biggest area.

All of these steps are shown at Figure 3.

After the pre-processing, one will have a unique filled circle, that is indeed the viewer age range smart sign.

## 3. CIRCLE DETECTION METHODS

The next step is the circle detection, which can be performed by many methods in literature [4-9], however at this work two algorithms, namely CHT and CE are employed.

### 3.1. CHT

There are many applications of CHT, such as traffic sign recognition [10] and analysis of geophysical data [11], which is indeed a special form of Generalized Hough Transform in pattern recognition [12].

As is well known, a circle with a radius of $r_0$, whose center is located at $(a_0,b_o)$ point on cartesian coordinates can be defined by the following equation:

$$(x_i - a_0)^2 + (y_i - b_0)^2 = r_0^2 \tag{1}$$

In the mapping from the $(x,y)$ geometric space to $(a,b)$ parameter space for a circle with a known radius $r_0$; the $(x_i,y_i)$ points taken over the circle, will generate new circles with the same radii, however centered on the mentioned points $(x_i,y_i)$. It is clear that, the number of circles formed on the parameter space will be equal to the number of points taken from the geometric space. All circles on the parameter space will exactly intersect at the $(a_0,b_0)$ center point of geometric space. This process is shown in Figure 4.

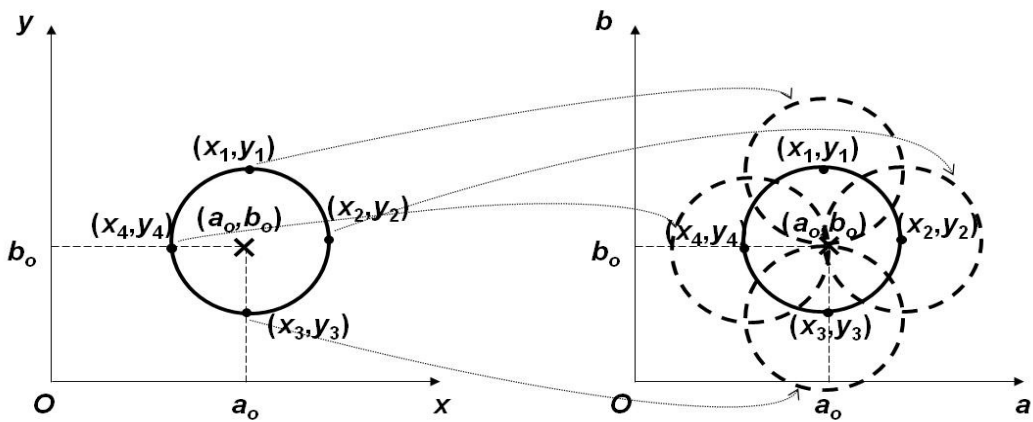

Figure 4. Mapping of a circle from geometric space to parameter space.

When the radius *r* is unknown, the problem becomes the determination of ($a_0$, $b_0$, $r_0$), which means a mapping from 2D geometric space to a 3D parameter space. Here, the radius for each point taken over the circle in geometric space varied from 0 to a finite value will form a cone on parameter space as shown in Figure 5. After all the points are chosen, the level where whole cones intersect with each other will give the seeked radius $r_0$.

CHT is the most popular method to identify circular objects at an image. However, despite its robustness, simplicity and performance, it requires computational burden, thus preventing the effective operation in real-time applications. For this reason, some other circle detection algorithms need to be used.

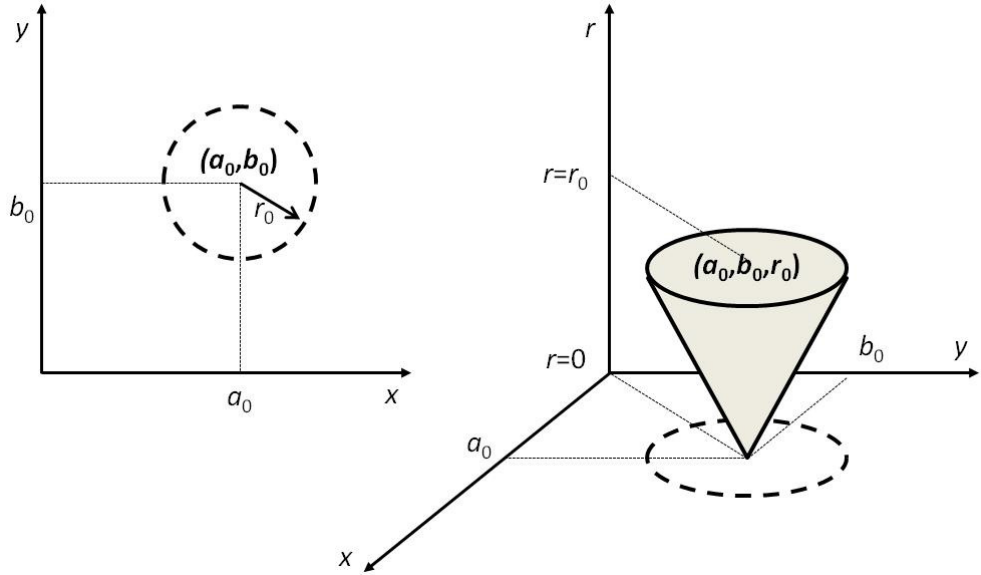

Figure 5. Mapping from a 2D geometric space to a 3D parameter space to determine the unknown radius $r_0$.

## 3.2. CE

A technique based on Least Squares Minimization (LSM) is performed to estimate the three unknown parameters ($a_0$, $b_0$, $r_0$), which are the center coordinates and radius.

According to LSM, the error function defined in (2) must be minimum for *n* points chosen from the filled circle after pre-processing.

$$E = \sum_{i=1}^{n} \left[ (x_i - a_0)^2 + (y_i - b_0)^2 - r_0^2 \right]^2 \quad (2)$$

For this purpose, the partial derivatives of the mentioned error function *E* with respect to parameters $a_0$, $b_0$, $r_0$ must be equal to zero at the same time.

$$\frac{\partial E}{\partial a_0} = 0; \frac{\partial E}{\partial b_0} = 0; \frac{\partial E}{\partial r_0} = 0 \quad (3)$$

The solution of the following matrix equation will realize this condition, providing the unknown parameters of the circle [4]:

Signal & Image Processing : An International Journal (SIPIJ) Vol.3, No.4, August 2012

$$\begin{bmatrix} \sum_{i=1}^{n}(x_i^3 + x_i y_i^2) \\ \sum_{i=1}^{n}(x_i^2 y_i + y_i^3) \\ \sum_{i=1}^{n}(x_i^2 + y_i^2) \end{bmatrix} = \begin{bmatrix} 2\sum_{i=1}^{n} x_i^2 & 2\sum_{i=1}^{n} x_i y_i & -\sum_{i=1}^{n} x_i \\ 2\sum_{i=1}^{n} x_i y_i & 2\sum_{i=1}^{n} y_i^2 & -\sum_{i=1}^{n} y_i \\ 2\sum_{i=1}^{n} x_i & 2\sum_{i=1}^{n} y_i & -n \end{bmatrix} \cdot \begin{bmatrix} \hat{a}_0 \\ \hat{b}_0 \\ \hat{z} \end{bmatrix} \quad (4a)$$

Where,

$$z = x_0^2 + y_0^2 - r_0^2 \quad (4b)$$

The method indeed, estimates the centers of gravities of the objects. It can assign a center point for every closed object other than a circle as shown in Figure 6. This problem is overcome by the classification process.

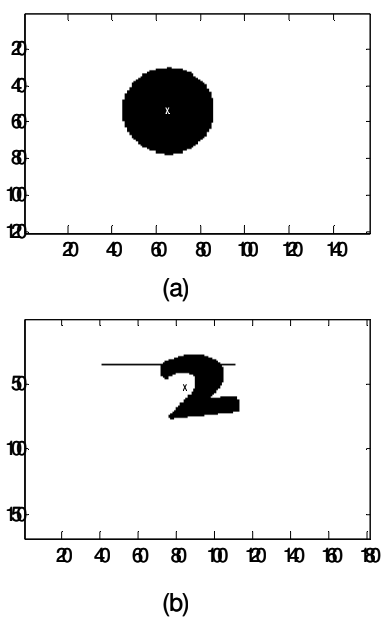

(a)

(b)

Figure 6. (a) The estimated parameters of the circle at the left-top, (b) The estimated parameters (Center of gravity) of the other closed object at the right-top of the image in figure 3a.

## 4. CLASSIFICATION

The classification process includes 2 steps:
1-) Feature extraction,
2-) Classification with an MLP.

### 4.1. Feature Extraction

After determining the center of the circular viewer age range smart sign, the pattern must be transformed to a feature extraction vector to introduce it to the MLP. This process is realized as follows: First, each sample is cropped with a range of its radius in height, and with a range of half of its radius in width from the center, in order to eliminate "+" and "1" characters of "+13" and "+18" symbols. Secondly, that rectangular form is thinned and resized as a 80x40 matrix. Afterwards, the number of black pixels, that lie at the left hand side of first white pixels at each row, are recorded. A new 80x1 vector, whose components are the counts of those black pixels, is formed and employed as the input of the MLP for the classification. All of these are shown in Figure 7.

However, some smart signs are exactly the negatives of the above. Therefore for such samples, all black pixels must be converted to whites and vice versa to apply the same procedure.

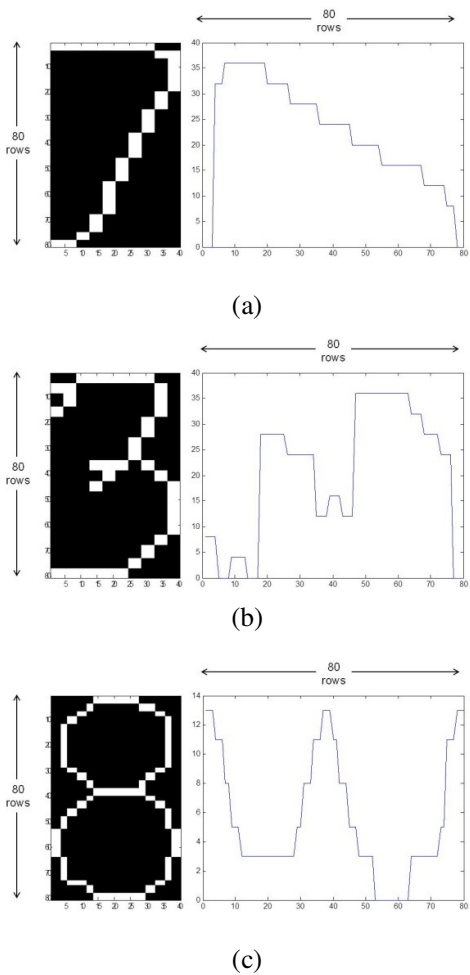

(a)

(b)

(c)

Figure 7. Generation of the feature extraction vectors for viewer age range smart signs:
(a) For 7+, , (b) For 13+, (c) For 18+

### 4.2. Classification with MLP

Neural networks (NNs) were inspired by the power, flexibility and robustness of the biological brain. NNs consist of many simple mathematical elements that work together in parallel and in series. Each neuron at this system has many inputs and only one output, and this output is an input to other neurons. A neuron model consists of a summing junction and an activation function. In this model, the output equation can be given as the following:

$$y = f(\sum_{i=1}^{n} w_i x_i + b) \tag{5}$$

Here, $x_i$s are inputs; $w_i$s are weight coefficients; $b$ is the bias, $y$ is the output and $f(x)$ is the activation function. A training process can be viewed as a problem of determining the network architecture and weight coefficients so that the neural network can perform a special task. A NN can adapt itself to get the desired response. NNs are trained from sample data, instead of programming by the inputs and the responses introduced to the system. For each input, the desired and obtained responses are compared, and the weight coefficients are changed accordingly to minimize the difference between the two. After an acceptable error is obtained, the learning process is stopped [13].

MLP is one of the most popular NN structures. In our work, the feature extraction vectors are applied to the input of a MLP having 15 cells in its hidden layer. Backpropogation (BP) algorithm is run for 18 samples from each viewer age range smart sign category at the training phase.

The outputs are classified as follows:
1-) 7+: No one under 7 admitted,
2-) 13+: No one under 13 admitted,
3-) 18+: No one under 18 admitted,
4-) N/C (Non/Classified=Ignored): All other patterns (e.g. the cropped porsion of the frame which doesn't include the viewer age range smart sign, as depicted before!).

The MLP structure and error approach are sketched in Figures 8 and 9, respectively.

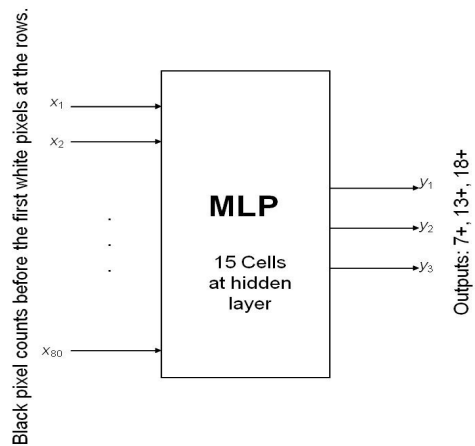

Figure 8. The MLP structure for classification.

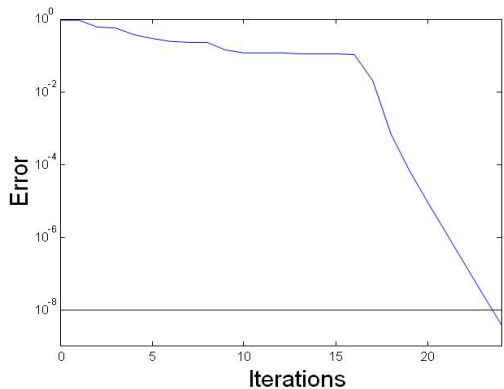

Figure 9. Error approach in the training with BP.

## 5. SIMULATIONS

Performances of CHT and CE methods are compared in detecting the viewer age range smart signs for the followings:

1-) Still images,
2-) Real time videos.

### 5.1. Simulations for Still Images

Since the digital TV broadcast consists of sequential frames, the performances of the detection processes including the CHT and CE methods on the still images play a key importance for the real time applications.

The determination of the center of the viewer age range smart sign at a frame taken from the broadcast with CHT and CE methods are given in Figures 10a and 10b, respectively.

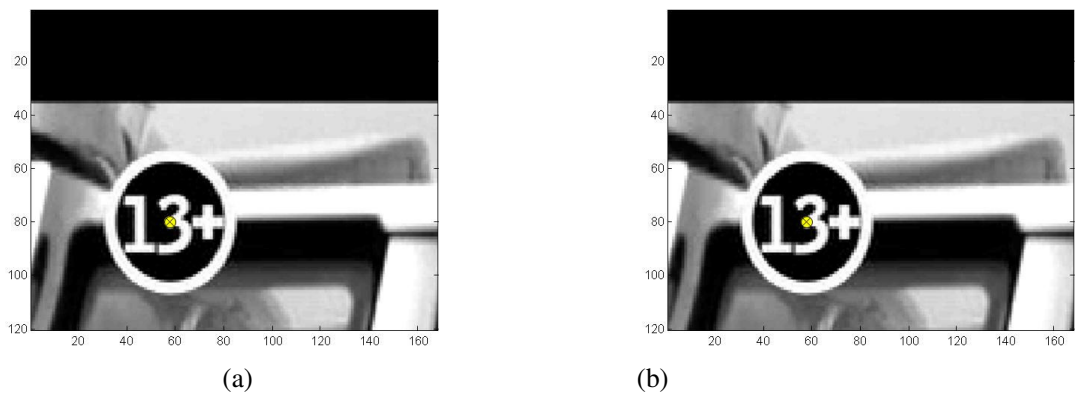

(a)      (b)

Figure 10. Determination of the center of a viewer age range smart sign:
(a) By using CHT, (b) By using CE.

Table 1 shows the summary of performances of the processes including CHT and CE for still images. Various samples are taken into account for each viewer age range smart sign (7+, 13+, 18+) and their means and standard deviations of the processing times at a standard PC are calculated. For example, 43 different samples for 7+ sign are processed both for CHT and CE.

(a) Performance of CHT.

| Sign | Sample # | Time-Delay (s) Mean | Time-Delay (s) Std.Dev. | Accuracy (%) |
|---|---|---|---|---|
| 7+ | 43 | 50.43 | 17.14 | 100 |
| 13+ | 27 | 39.10 | 16.11 | 100 |
| 18+ | 41 | 51.59 | 19.34 | 97.56 |

(b) Performance of CE.

| Sign | Sample # | Time-Delay (s) Mean | Time-Delay (s) Std.Dev. | Accuracy (%) |
|---|---|---|---|---|
| 7+ | 43 | 0.22 | 0.07 | 97.67 |
| 13+ | 27 | 0.23 | 0.06 | 100 |
| 18+ | 41 | 0.22 | 0.02 | 92.68 |

### 5.2. Simulations for Real Time Videos

As seen from Figure 11, the frame samples are picked from the broadcast, in order to realize the detection and classification processes. Since the viewer age range smart signs appear approximately for 10s, a sampling period of 4s will be sufficient, however all the steps given in Figure 2 have to be completed within this interval before the next sample is taken.

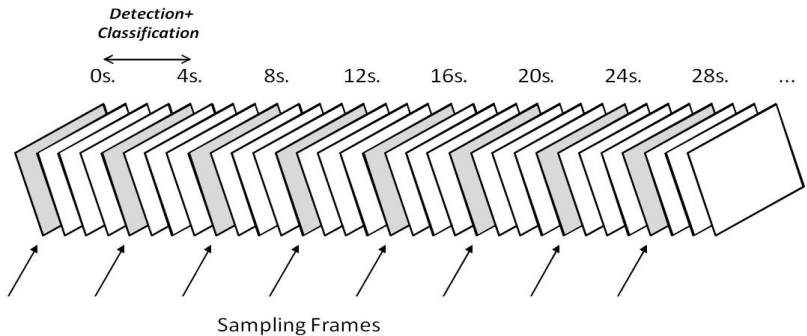

Figure 11. Picking the sample frames from the digital broadcast.

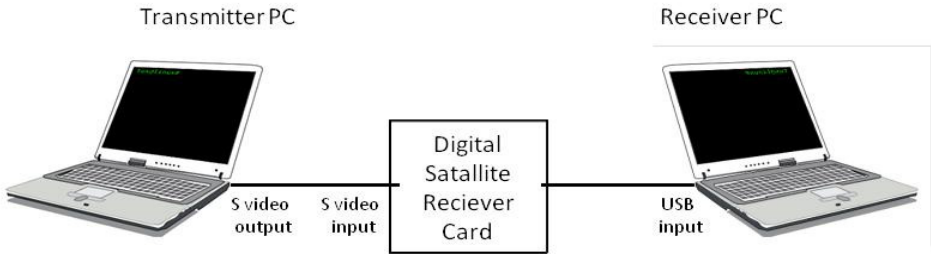

Figure 12.Experimental setup for the real time detection and classification process.

All these are relevant to the speed of the algorithm. Therefore according to Table 1, the process including CE will be much more suitable when compared to the one that employs CHT.

At the simulations for real time operations, two standard PCs - one is employed as the transmitter which broadcasts 720x576 MPEG4 videos and the other as the receiver – are used. That experimental setup is given in Figure 12.

The pre-processing, CE – not the CHT - and the classification steps are carried out with a software run at the receiver PC for the frames taken from the broadcast and an output image seen in Figure 13 is generated which gives the information about the viewer age range sign and its location.

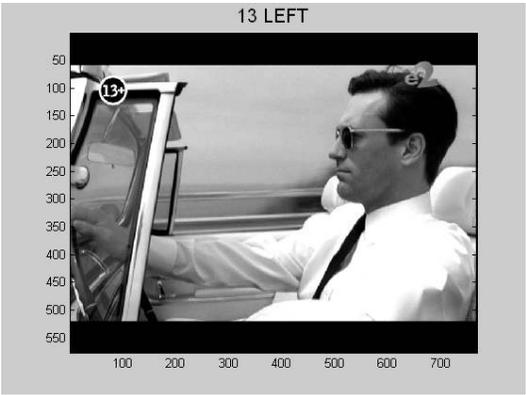

Figure 13.Output for the viewer age range smart sign and its location.

## 6. CONCLUSIONS

In this paper, the viewer age range smart signs at the broadcasts of Turkish TV channels are automatically detected and classified.For this purpose, two popular circle segmentation methods namely, CHT and CE are employed. However primarily, a pre-processing is carried out in order to reduce the computational burden, which includes basic image processing procedures such as cropping, edge detection, filling etc.Both still images and streaming video are processed by a program run at a standard PC and satisfactory performances are observed.

It is believed that, the written software is suitable for real time applications, leading the design of TV receivers which are sensible to the mentioned signs. As future works, the performances of other circle detection algorithms can be tested and/or an hardware can be realized as an embedded system as well as by using digital signal processing cards [14].

## ACKNOWLEDGEMENTS

The authors would like to acknowledge to Research Fund of Kadir Has University who supplied all of the hardware at this work.

## REFERENCES

[1]   http://www.rtukisaretler.gov.tr/RTUK/index.jsp.
[2]   http://www.mpaa.org/ratings/what-each-rating-means.
[3]   J.S. Lim, "Two Dimensional Signal and Image Processing", Prentice-Hall, 1989.
[4]   W. Yi, "A Fast Finding and Fitting Algorithm to Detect Circles", in Proceedings of IGARSS'98: IEEE Geoscience and Remote Sensing Symposium, Seatttle, USA, pp.1187-1189, 1998.
[5]   Kelvin S.Y. Yuen and Eric K.H. Lo, "A Course-to-Fine Approach for Circle Detection", in Proceedings of ISSIPNN'94: International Symposium on Speech, Image Processing and Neural Networks, Hong Kong, pp.523-526, 1994.
[6]   T.J. Atherton and D.J. Kerbyson, "Size Invariant Circle Detection", Elsevier Image and Vision Computing, vol.17, no.11, pp.795-803, 1999.
[7]   W. Cai, Q. Yu and H. Wang, "A Fast Contour-Based Approach to Circle and Elipse Detection", in Proceedings of WCICA'04: 5th World Congress on Intelligent Control, Hangzhou, China, pp.4686-4690, 2004.
[8]   G.M. Schuster and A.K. Katsaggelos, "Robust Circle Detection Using a Weighted MSE Estimator", in Proceedings of ICIP'04: IEEE International Conference on Image Processing, Singapore, pp.2111-2114, 2004.
[9]   B. Lamiroy, O. Gaucher and L.Fritz, "Robust Circle Detection", in Proceedings of ICDAR'07: IEEE 9th International Conference on Document Analysis and Recognition, Curituba, Brasil, pp.526-530, 2007.
[10]  Y.B. Damavandi and K. Mohammadi, "Speed Limit Traffic Sign Detection and Recognition", in Proceedings of CIS'04: IEEE Confererence on Cybernetics and Intelligent Systems, Singapore, pp.797-802, 2004.
[11]  G.R.J. Cooper and D.R. Cowan, "The Detection of Circular Features on Irregularly Spaced Data", Elsevier Computers and Geosciences, vol.30, no.1, pp.101-105, 2004.
[12]  D.H. Ballard, "Generalizing the Hough Transform to Detect Arbitrary Shapes", Elsevier Pattern Recognition,vol.13, no.2, pp.111-122, 1981.
[13]  S. Baktır,et. al.,"Detection of Trojans in Integrated Circuits",in Proceedings of INISTA 2012: International Symposium on Innovations in Intelligent SysTems and Applications, July 2012, Trabzon.
[14]  "Embedded Processing Guide", Texas Instruments 3Q, 2009.